\documentclass[10pt, a4paper]{article}
\usepackage[T1]{fontenc}
\usepackage[utf8]{inputenc}

\usepackage[final]{lrec2026} 
\usepackage{booktabs}

\usepackage{listings}
\usepackage[table]{xcolor}  
\definecolor{kgrow}{gray}{0.9}      

\definecolor{listingbg}{RGB}{248,248,248} 
\definecolor{listingrule}{RGB}{200,200,200}
\usepackage{amsmath} 
\usepackage{siunitx} 
\usepackage{amssymb} 
\usepackage{enumitem} 

\title{Do LLMs Know What Luxembourgish Borrows? Probing Lexical Neology in Low-Resource Multilingual Models}

\name{
  Nina Hosseini-Kivanani$^{* 1,2}$}

\address{$^{1}$ University of Luxembourg, Luxembourg\\
         $^{2}$ Radio Télévision Luxembourg (RTL), Luxembourg\\
         {\small nina.hosseinikivanani@ext.uni.lu}}

\abstract{
Large language models (LLMs) are increasingly used for writing assistance in small contact languages, yet it is unclear whether they respect community norms around lexical borrowing and neology. We introduce LexNeo-Bench, a 3{,}050-instance token-level benchmark derived from LuxBorrow, a large-scale Luxembourgish news corpus, where target tokens are labelled as native or as French, German, or English borrowings. Using this benchmark, we probe three multilingual LLMs across 34 prompt settings on two tasks: borrowing type classification and a binary lexical-innovation proxy (borrowing versus native). Without external context, models perform only slightly above chance on borrowing classification, so we construct a linguistic knowledge graph that encodes donor language, morphological patterns, and lexical analogues, and inject instance-specific subgraphs into the prompt. Knowledge-graph prompts raise borrowing classification accuracy from 25 -- 35\% up to 71 -- 81\%  and largely close the gap between small and large models, while leaving neology detection difficult and sensitive to few-shot design. Our results show that lexicon-aware prompting is highly beneficial for robust borrowing judgments in low-resource contact languages and that lexical resources can serve as structured context for LLM evaluation. This study was carried out within the ENEOLI COST Action and examines borrowing as a form of lexical innovation in multilingual Luxembourgish data.
 \\ \newline  }

\begin{document}

\maketitleabstract

\section{Introduction}
Neology, the creation and diffusion of new lexical items, has long been central to lexicography, corpus linguistics, and sociolinguistics. With the emergence of large language models (LLMs), neology enters a new phase. LLMs are trained on massive multilingual corpora, absorb existing neologisms, and can themselves generate novel forms, blends, and hybrid structures in response to prompts. This raises questions not only about how LLMs detect and represent lexical innovation, but also about how their behavior interacts with existing norms and resources in individual language communities \cite{wolfer2023tracking,zheng2024neo}.

For smaller languages such as Luxembourgish, lexical innovation is tightly intertwined with contact phenomena. Luxembourgish exists in a dense contact zone with German, French, and English. Much of its modern lexical growth is realized through borrowing and adaptation from these donor languages rather than through entirely endogenous coinages \cite{adda2008developments}. In written media, especially professionally edited news, many emergent forms are morphologically or orthographically integrated into Luxembourgish, while others remain closer to code-switching \cite{lavergne2014automatic}. For downstream Natural Language Processing (NLP) tools and LLM-powered applications, it matters whether these items are recognized as legitimate Luxembourgish words or treated as errors, foreign insertions, or targets for normalization back to French or German.

Previous work on Luxembourgish borrowing introduced LuxBorrow~\cite{hosseini-kivanani2026luxborrow}, a large-scale corpus of Radio Télévision Luxembourg (RTL) news (1999--2025) annotated with sentence-level language identification and token-level labels for native items, borrowings from French, German, and English, and code-switching. That study focused on contact linguistic patterns and diachrony, showing that Luxembourgish remains the matrix language in news, while lexical borrowing and code mixing are pervasive but low-intensity, with a rich inventory of morphological and orthographic adaptation patterns. However, LuxBorrow did not address how contemporary LLMs treat these adapted forms, nor whether they recognize them as part of the Luxembourgish lexicon.

In this paper, we treat morphologically and orthographically adapted borrowings in Luxembourgish news as a key locus of lexical innovation and use LuxBorrow as ground truth to evaluate neology awareness in multilingual LLMs. We construct a token-level classification benchmark that pairs Luxembourgish sentences from RTL.lu with highlighted target tokens and gold labels indicating whether each token is native or a borrowing, and if so, from which donor language. On top of this benchmark, we define two tasks: a borrowing classification task in which models choose from four labels (NATIVE, FR\_LOAN, DE\_LOAN, and EN\_LOAN
as a diagnostic distractor) but are evaluated on three gold classes, and a binary neology decision task.

Our study is organized around three research questions.

\begin{itemize}
  \item RQ1. To what extent do off-the-shelf multilingual LLMs correctly classify native Luxembourgish words and distinguish French- vs German-origin adapted borrowings in RTL news?
  \item RQ2. Do LLMs systematically bias their judgments toward dominant donor languages, especially French and German, and how often do they incorrectly project English-origin hypotheses via the EN\_LOAN distractor label?
  \item RQ3. How does providing explicit lexicon-based context, for example, a loanword registry, affect LLM performance and their treatment of Luxembourgish lexical innovation?
\end{itemize}

To answer these questions, we evaluate three strong multilingual LLMs in frozen, prompt-only mode. We compare zero-shot prompting, few-shot prompting with manually chosen examples of Luxembourgish borrowings, and two knowledge-based prompting conditions: \texttt{KG\_flat}, which provides a global list of borrowing patterns, and \texttt{KG\_graph}, which injects an instance-specific lexicon context derived from the LuxBorrow loanword registry.



Our contributions are threefold. First, we introduce LexNeo-Bench, a token-level benchmark for borrowing classification in Luxembourgish, derived from LuxBorrow, with public scripts for extraction, prompting, and evaluation. Second, we add a binary lexical-innovation proxy task that collapses borrowings versus native items to probe neology awareness, and show that it remains challenging even for strong multilingual LLMs. Third, we show that lightweight lexicon-based context via a linguistic knowledge graph can substantially improve borrowing judgments in a low-resource contact language, which suggests concrete avenues for integrating community-curated lexical resources into LLM prompting for neology-sensitive applications. Within the ENEOLI COST Action, this study contributes to WG2 by treating borrowing in Luxembourgish as a corpus-based case of lexical innovation and by evaluating how multilingual LLMs analyze such forms in a low-resource contact setting.

\section{Related Work}

\subsection{Borrowing, code-switching, and neology}

Contact linguistics distinguishes lexical borrowing, items integrated into the recipient language’s lexicon and grammar, from code-switching, that is, spontaneous alternation between languages within discourse. Classic accounts emphasize that entrenched borrowings are morphologically and phonologically integrated, frequent, and often listed in dictionaries, while code-switches retain donor language structure and remain more speaker-specific. This view underlies the ``Simple View'' of borrowing, which operationalizes the difference in terms of listedness in the mental lexicon and community entrenchment~\cite{treffersdaller2025simple,chesley2010predicting}.

In multilingual European contexts, written media often show a stable matrix language with pervasive but shallow insertions from donor languages. Borrowings can be introduced via institutional domains such as politics, finance, and administration, before diffusing into more general registers. Over time, morphologically adapted forms may compete with native synonyms or with less integrated loan variants. This dynamic is particularly visible in Luxembourgish, where French and German both supply a rich inventory of technical and everyday lexical items, and where orthographic and morphological adaptation blur the surface boundary between native and borrowed forms~\cite{anastasiou2022deliverable,lavergne2014automatic,adda2008developments}.

Lexicographic and corpus-based studies of neology therefore give prominence to borrowed and adapted items when tracking lexical innovation, especially in small languages that rely heavily on lexical importation from regional lingua francas~\cite{wolfer2023tracking}.

\subsection{Computational borrowing and neology detection}

In NLP, early work on multilingual text mixing emphasized document- or
utterance-level indices, such as code-mixing indices and entropy-based
measures, which treat all foreign tokens uniformly. More recent studies
move to explicit borrowing detection and distinguish unassimilated
foreign tokens, code-switches, and integrated loanwords. This line of work
has introduced borrowing-annotated corpora, for example anglicism
detection in Spanish newswire~\cite{alvarezmellado2020annotated,alvarez2021extracting},
and shared tasks with sequence tagging baselines~\cite{adobo2021,adobo2025}.
Methods range from conditional random fields and BiLSTM-CRFs to transformer
taggers that incorporate lexical and orthographic
features~\cite{alvarezmellado2020annotated,alvarezmellado2020lazaro},
alongside resource-lean approaches to code-switching identification that
rely mainly on word lists and monolingual corpora~\cite{kevers2022coswid}.

Beyond borrowing per se, neology detection has traditionally relied on
dictionary versus corpus comparisons combined with temporal information,
for example, locating forms that appear in recent corpora but are absent
from older lexica. With the advent of LLMs, recent work has begun to
integrate these models into neologism detection pipelines, for example
using them as filters or validators for candidate neologisms, and to
provide lemmata and definitions for emergent forms~\cite{tomaszewska2025neon,hosseini2025hybrid}.
Other studies highlight how LLMs can also generate non-attested
``LLM neologisms'' due to tokenization and encoding artifacts~\cite{iwamoto2024llm}.
This opens a new evaluation axis: not just whether LLMs can help detect
neology, but whether their intrinsic lexical knowledge and biases align
with community norms and lexicographic resources~\cite{tomaszewska2025neon,hosseini2025hybrid,iwamoto2024llm}.

\subsection{LLMs, low resource languages, and lexical inequality}
Work on LLMs in low-resource languages highlights skewed coverage and performance gaps, where models trained mainly on high-resource languages underrepresent or mis-analyze items from smaller languages and can ``normalize'' adapted borrowings back to donor forms. This has consequences for spell-checkers, assistive writing tools, and generation systems that interact with speakers of contact languages. Empirical studies of Luxembourgish resources and their multilingual context document sparse written production and heavy code-mixing and adaptation pressure, which exacerbate LLM coverage problems~\cite{plum2024luxbank,lavergne2014automatic,adda2008developments}.

Lexicon-aware prompting and retrieval/gazetteer augmentation show that injecting compact community resources into LLM workflows yields large gains on complex Named Entity Recognition (NER) and entity-centric tasks~\cite{tan2023damo,chen2022ustc}. This motivates using curated loanword registries or structured knowledge-graph hints to probe LLM judgments about borrowed and adapted forms. Against this background, we use a borrowing-annotated Luxembourgish news corpus as a neology resource to build LexNeo-Bench, a benchmark that probes LLM lexical decisions in a dense contact setting.


\section{Experiments}
\label{sec:experiments}
\begin{figure}[hpt!]
    \centering
    \includegraphics[width=\linewidth]{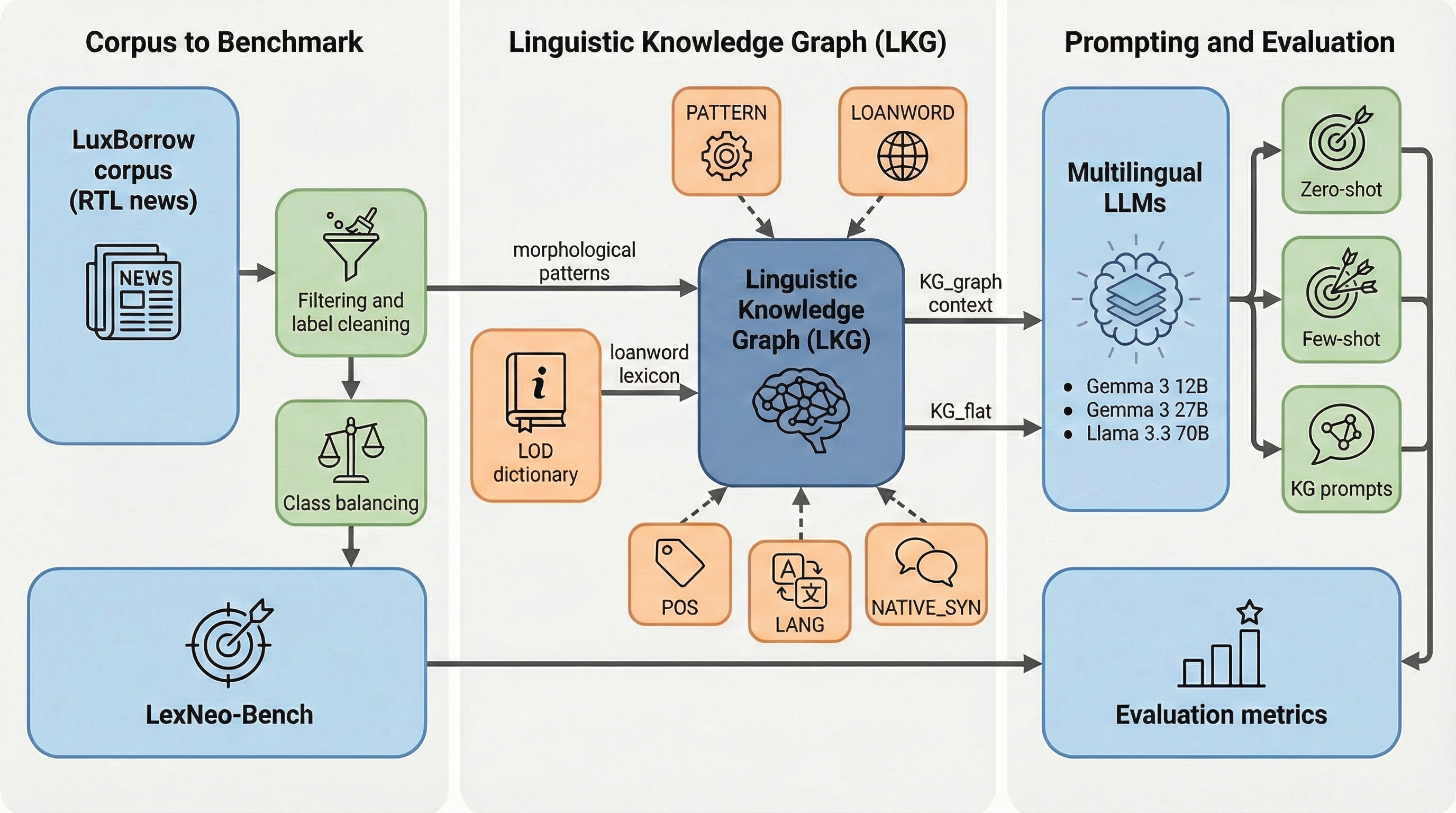}
    \caption{LexNeo-Bench pipeline.} 
    \label{fig:pipeline}
\end{figure}

Figure~\ref{fig:pipeline} summarizes the overall evaluation pipeline, from LuxBorrow-derived benchmark construction and LKG retrieval to prompt assembly and multilingual LLM evaluation.

\subsection{Benchmark construction}
\label{subsec:benchmark}

We construct LexNeo-Bench, a token-level evaluation benchmark derived from the LuxBorrow corpus of professionally edited Luxembourgish news. LuxBorrow provides sentence-level language identification over RTL articles together with token-level borrowing labels generated by a morphological pattern pipeline. Each benchmark instance consists of a Luxembourgish sentence, a highlighted target token, its gold borrowing label, compact morphological evidence, and article metadata such as section and timestamp.

The label space follows the LuxBorrow taxonomy and distinguishes native Luxembourgish items (\textbf{NATIVE}) from French-, German-, and English-origin borrowings (\textbf{FR\_LOAN}, \textbf{DE\_LOAN}, \textbf{EN\_LOAN}). Tokens tagged as \textbf{CODE\_SWITCH} or as named entities are excluded from the main task, since the focus is on entrenched lexical items rather than span-level alternation. To avoid extremely sparse classes, we discard labels with fewer than 50 instances in the source corpus. EN\_LOAN appears only 24 times and is therefore removed from the evaluation set, which yields a three-way task over NATIVE, FR\_LOAN, and DE\_LOAN, even though the conceptual four-way taxonomy is kept in the prompts.

The original LuxBorrow corpus comprises \num{259305} RTL news articles and 43.7 million tokens. We first remove punctuation, a curated list of Luxembourgish function words, and tokens with low-confidence automatic labels. From the remaining pool, we draw \num{1000} instances per active class, which results in a balanced benchmark of 3\,000 examples, and we add a small diagnostic stratum of 50 CODE\_SWITCH tokens for error analysis. The final benchmark therefore contains \num{3050} instances:
\(\text{NATIVE}=1000\), \(\text{FR\_LOAN}=1000\), \(\text{DE\_LOAN}=1000\), and \(\text{CODE\_SWITCH}=50\).

Each instance inherits the publication date of its source article (1999--2025), providing a diachronic signal. As a proxy for entrenchment, we contrast tokens from articles published before 2015 with those after 2015. This split is not first-attestation dating, but leverages a 25-year professionally edited news record. The borrowing labels serve as the primary gold standard for both tasks. For the auxiliary neology decision task, we derive a binary label at prompt construction time by mapping tokens annotated as \texttt{FR\_LOAN}, \texttt{DE\_LOAN}, or \texttt{EN\_LOAN} to \texttt{YES} (lexical innovation) and \texttt{NATIVE} tokens to \texttt{NO}. The recent versus established flag is used only for temporal robustness analysis.


For prompt types that embed full lexicon entries, we restrict ourselves to a reduced subset of 674 benchmark items for which dictionary definitions, etymology, and related lexical information are available in a consistent format from the Lëtzebuerger Online Dictionnaire (LOD)~\citelanguageresource{LOD}. This avoids noisy or incomplete context in lexicon-assisted conditions.

\subsection{Linguistic Knowledge Graph}
\label{subsec:lkg}


To provide models with structured linguistic context, we construct a \textbf{Linguistic Knowledge Graph (LKG)} that integrates three LuxBorrow-related resources: a compiled index of productive morphological patterns, a loanword lexicon extracted from LOD, and a hand-curated table of loanword–native synonym pairs.

Nodes are typed as morphological patterns (PATTERN), donor languages (LANG), loanword entries (LOANWORD), native Luxembourgish synonyms (NATIVE\_SYN), and part-of-speech tags (POS). Edges encode linguistic relations such as pattern membership (\textit{follows\_pattern}), donor origin (\textit{from\_donor}), synonymy (\textit{has\_synonym}), competition between patterns with the same Luxembourgish affix but different donor languages (\textit{contrastive}), and lexical category (\textit{has\_pos}).

For each benchmark instance, we perform multi-hop retrieval on this graph to construct a compact, token-specific explanation subgraph. Starting from the target token, we match compatible LuxBorrow patterns, query LOD to obtain donor metadata, and reconstruct an etymology-style chain of the form
\(\text{donor form} \rightarrow \text{adaptation pattern} \rightarrow \text{Luxembourgish form}\),
collect linked native synonyms, and sample a small set of analogues that share a pattern, plus a few contrastive patterns with the same affix but different donor languages. The retrieved subgraph is then linearized into a structured natural-language block of at most 30 lines and prepended to the model prompt. This instance-specific \texttt{KG\_graph} context replaces a much coarser \texttt{KG\_flat} baseline in which the same global list of 19 patterns is appended to every example independently of the target token.

\subsection{Prompt setups}
\label{subsec:prompts}

All prompts share a common two-role template. The system message defines a Luxembourgish linguistics expert persona, and the user message concatenates optional knowledge-graph context, the task instruction, and the Luxembourgish sentence with the target token marked by \texttt{**}. 
The user message then introduces any external context, followed by a concise instruction to assign exactly one label to the highlighted token and the sentence in which it appears.

We evaluate a family of prompt strategies for both borrowing classification and neology detection. In all cases, the model receives a short English instruction, the Luxembourgish sentence, and the target token marked with \texttt{**}. For the \textbf{classification} task, the model must output exactly one label from the conceptual four-way set \(\{\texttt{NATIVE}, \texttt{FR\_LOAN}, \texttt{DE\_LOAN}, \texttt{EN\_LOAN}\}\). Although \texttt{EN\_LOAN} does not appear in the evaluation data, we keep it as a possible answer to capture uncertainty toward English-origin candidates. For the neology task, the model must answer \texttt{YES} if the token should be treated as a lexical innovation in Luxembourgish, and \texttt{NO} otherwise.

The base prompt strategies include a plain \texttt{zero\_shot} condition (system role plus task description, no additional context), a \texttt{few\_shot} variant with five manually authored demonstrations, and a \texttt{minimal} variant that reduces the instruction to a single line and enforces label-only output. The five demonstrations do not overlap with the 3\,050 benchmark instances. They consist of two prototypical \texttt{FR\_LOAN} examples and one example each for \texttt{DE\_LOAN}, \texttt{EN\_LOAN}, and \texttt{NATIVE}. Each demonstration pairs a short Luxembourgish sentence with its gold label and a brief linguistically motivated justification.

An excerpt of the few-shot prompts is shown in Listing~\ref{lst:prompts}. 

\begin{lstlisting}[
  caption={Excerpt of few-shot prompts for borrowing classification and neology decision.},
  label={lst:prompts},
  basicstyle=\ttfamily\scriptsize,
  aboveskip=0.3em,
  belowskip=0.3em
]
System:
  You are a linguistic expert specializing in Luxembourgish
  (Lëtzebuergesch). Luxembourgish is a West Germanic language spoken
  in Luxembourg that regularly borrows and morphologically adapts
  words from French, German, and English.

User (classification task):
  Given the following Luxembourgish sentence and the highlighted
  word (marked with ** **), decide whether the highlighted word is:
    - NATIVE:  a native Luxembourgish word
    - FR_LOAN: a borrowing from French (morphologically adapted
               into Luxembourgish)
    - DE_LOAN: a borrowing from German (morphologically adapted
               into Luxembourgish)
    - EN_LOAN: a borrowing from English (morphologically adapted
               into Luxembourgish)
  Respond with ONLY the label on the first line and a one-sentence
  justification on the second line.

  Example:
    Sentence: D'**Pompjeeën** hunn de Brand schnell ënnert
              Kontroll bruecht.
    Assistant: FR_LOAN
    Justification: 'Pompjeeën' derives from French 'pompier',
      adapted with the Luxembourgish plural suffix "-en" and
      spelling "ee" for /e:/.

User (neology task):
  Given the following Luxembourgish sentence and the highlighted
  word, decide whether this token should be treated as a lexical
  innovation in Luxembourgish.
  Answer YES or NO, followed by one sentence of explanation.

  Sentence: [Luxembourgish sentence containing **TOKEN**].
\end{lstlisting}

Knowledge-augmented prompts add morphological information derived from LuxBorrow and LOD. In the \texttt{KG-flat} conditions, the user message begins with a preamble
\emph{``According to the LOD, the following morphological adaptation patterns are productive in Luxembourgish:''}, followed by up to twenty globally fixed pattern entries. Each entry lists a pattern name, its type (morphological, orthographic, or lexical), the donor language, and up to three example pairs, for example \emph{``\textit{éiere} $\rightarrow$ \textit{er}, type. morph, donor. FR, e.g. \textit{abordéieren} $\leftarrow$ \textit{aborder}''}. This global pattern block is identical in \texttt{KG-flat} and is appended to every instance, which contrasts with \texttt{KG-graph}, where the context is an instance-specific LKG subgraph as described in Section~\ref{subsec:lkg}.

To quantify the contribution of individual LKG components, we define six ablation variants that selectively remove lexicon attestation, etymology chains, synonym links, analogical examples, or contrastive patterns, as well as a \texttt{lex-only} condition that keeps only dictionary-style information without graph structure. 
Together, the eleven base strategies and six ablations define 17 prompt setups per task. Applied to both borrowing classification and neology detection, this yields 34 task-specific evaluation settings per model.

\subsection{Models}
\label{subsec:models}

All experiments are conducted with instruction-tuned, general-purpose LLMs accessed through an OpenAI-compatible endpoint (OpenRouter API). We deliberately treat the models as frozen black boxes and rely exclusively on prompting; no fine-tuning is performed.

We consider three model sizes: Gemma~3~12B (\texttt{google/gemma-3-12b-it}) as a small model, Gemma~3~27B (\texttt{google/gemma-3-27b-it}) as a medium model, and Llama~3.3~70B~Instruct (\texttt{meta-llama/llama-3.3-70b-instruct}) as a large model. All runs use a temperature of 0.0 and a maximum output length of 1{,}024 tokens to enforce deterministic, label-complete responses. Combining three models with 34 prompt configurations yields 102 evaluation settings, each applied to the full LexNeo-Bench of 3\,050 instances.

\subsection{Evaluation protocol}
\label{subsec:evaluation}
Model outputs often contain explanations or formatting artifacts, so we post-process responses to recover a single canonical label per instance. We strip explicit reasoning blocks (for example between \texttt{\textless think\textgreater} and \texttt{\textless/think\textgreater}), then examine the first and last non-empty lines and map them to one of the allowed labels using a small normalization dictionary (for example, \texttt{FRENCH}, \texttt{FR}, or \texttt{FR\_loanword} all map to \texttt{FR\_LOAN}, while \texttt{LUXEMBOURGISH} or \texttt{LB} map to \texttt{NATIVE}). Outputs that cannot be unambiguously resolved are marked as \texttt{PARSE\_ERROR} and omitted from metric computation; we report their frequency separately.

For the borrowing classification task, we report accuracy, balanced accuracy, macro- and weighted-F1 over the active classes, as well as per-class precision, recall, F1, and confusion matrices. In addition, we analyze two derived sub-tasks: a binary native versus borrowed decision (collapsing \texttt{FR\_LOAN} and \texttt{DE\_LOAN}) and donor-only discrimination between \texttt{FR\_LOAN} and \texttt{DE\_LOAN}. For the neology task, we treat \texttt{YES} as the positive label and report
accuracy, precision, recall, and F1, with additional breakdowns by donor
language. The gold label is derived directly from the primary LuxBorrow
borrowing annotation: tokens annotated as \texttt{FR\_LOAN}, \texttt{DE\_LOAN},
or \texttt{EN\_LOAN} are mapped to \texttt{YES} (lexical innovation), and
\texttt{NATIVE} tokens to \texttt{NO} (see Section~\ref{subsec:benchmark}).
All metrics are computed on the same fixed test set.


Temporal robustness is assessed by comparing accuracies on established versus recent items and reporting the absolute gap. Finally, the evaluation pipeline supports resumable execution. Prediction files are incrementally extended when experiments are restarted, which makes large grids of runs robust to interruptions without recomputation.

\section{Results}
\label{sec:results}

\subsection{RQ1. Borrowing classification performance}
\label{subsec:rq1}

Table~\ref{tab:overall_performance} summarizes three-way borrowing classification accuracy and macro F1 across models and prompt strategies. Without a structured linguistic context, performance remains modest.
In the zero-shot baseline, accuracy ranges from 24.5\% for
Gemma~3~12B to 34.7\% for Llama~3.3~70B, and more elaborate non-KG
prompts, such as Few-shot, remain below 42\%
across all models. 

Since models choose from four output labels, a random baseline
yields 25\% accuracy; zero-shot performance ranges from 24.5\% to
34.7\%, indicating that parametric knowledge alone barely exceeds
chance.

Introducing a structured linguistic context via the KG-graph condition
changes this picture sharply. With KG-graph, accuracy rises to 81.0\% for Gemma~3~12B, 71.4\% for Gemma~3~27B, and 71.3\% for Llama~3.3~70B, and macro F1 exceeds 0.55
for all models, peaking at 0.634 for Gemma~3~12B. The simpler KG-flat variant, which exposes only a global list of morphological patterns, does not close this gap and behaves similarly to non-KG baselines. The improvement, therefore, stems from instance-specific retrieval rather than merely reminding the model that borrowing patterns exist. Taken together, these results answer RQ1 by showing that structured, token-level linguistic context is necessary to achieve robust borrowing classification in Luxembourgish.

\begin{table}[hpt!]
  \centering
  \scriptsize
  \setlength{\tabcolsep}{7pt}
  \renewcommand{\arraystretch}{0.4}
  \caption{Acc.\ and macro F1 (in \%), and KG gain $\Delta_{\mathrm{KG}}$ (percentage
points), defined as the accuracy difference between KG-graph and zero-shot.}
  \label{tab:overall_performance}
  \begin{tabular}{l l r r r}
    \toprule
    \textbf{Model}         & \textbf{Prompt}    & \textbf{Acc.(\%)} & \textbf{Macro F1} & $\Delta_{\mathrm{KG}}$ \\
    \midrule
    Gemma~3~12B   & Zero-shot & 24.5 & 22.3     &        \\
                  & Few-shot  & 38.3 & 30.3     &        \\
                  & KG-flat   & 30.3 & 26.2     &        \\
    \rowcolor{kgrow}
                  & KG-graph  & 81.0 & 63.4     & +56.5  \\
    \midrule
    Gemma~3~27B   & Zero-shot & 31.4 & 19.7     &        \\
                  & Few-shot  & 38.0 & 29.8     &        \\
                  & KG-flat   & 33.7 & 22.6     &        \\
    \rowcolor{kgrow}
                  & KG-graph  & 71.4 & 55.9     & +40.1  \\
    \midrule
    Llama~3.3~70B & Zero-shot & 34.7 & 27.9     &        \\
                  & Few-shot  & 38.0 & 29.0     &        \\
                  & KG-flat   & 36.3 & 27.9     &        \\
    \rowcolor{kgrow}
                  & KG-graph  & 71.3 & 55.7     & +36.6  \\
    \bottomrule
  \end{tabular}
\end{table}

\subsection{RQ2. Per class performance and donor bias}
\label{subsec:rq2}

\begin{figure*}[hpt!]
  \centering
  \includegraphics[width=1\linewidth]{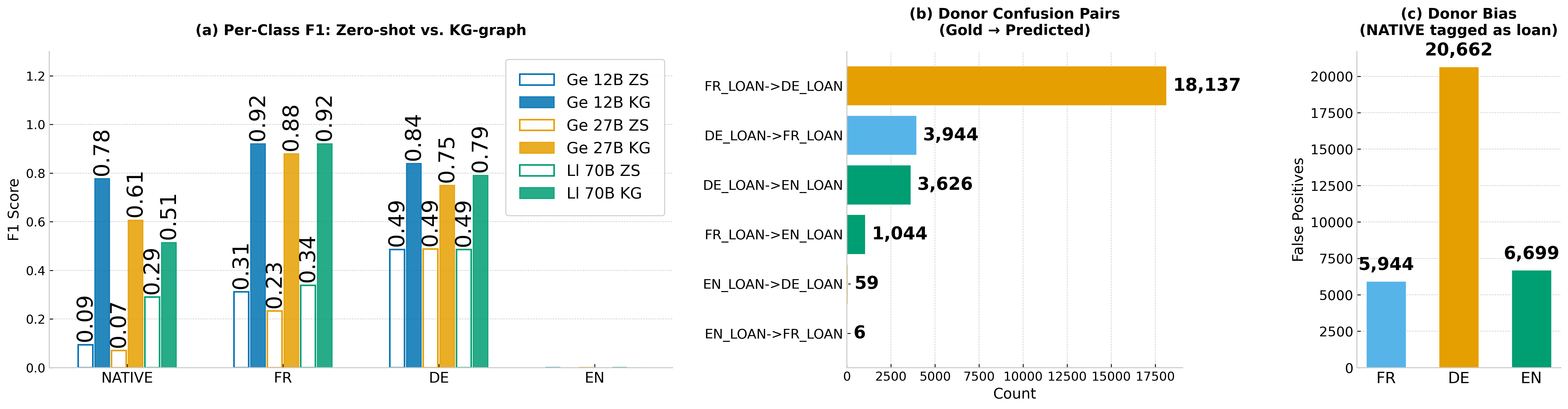}
  \caption{(a) Per-class F1 for zero-shot and KG-graph by model (GE12, GE27, \& LL70 denote Gemma~3~12B, Gemma~3~27B, \& Llama~3.3~70B, respectively.).
           (b) Top donor confusion pairs.
           (c) False-positive rates on \textsc{NATIVE}: proportion of \textsc{NATIVE} tokens predicted as loans.} 
  \label{fig:combined}
\end{figure*}

To understand where the gains from KG-graph conditioning arise, Figure~\ref{fig:combined}(A) reports per class F1 under the
KG-graph prompt. All three models achieve strong F1 scores for French and German
borrowings. Gemma~3~12B reaches 0.920 for FR\_LOAN (French borrowing) and 0.840 for
DE\_LOAN (German borrowing); Gemma~3~27B reaches 0.880 and 0.750 respectively, and Llama~3.3~70B scores 0.921 and 0.791. Performance on NATIVE items is more variable.
Gemma~3~12B maintains a solid 0.777 F1, whereas Llama~3.3~70B drops to
0.515, suggesting that the larger model overfits to donor cues and sometimes over-predicts borrowing for genuinely native words.



Confusion patterns show a marked donor asymmetry. In Figure~\ref{fig:combined}(b), the dominant error is French-origin items misclassified as German-origin (\textsc{FR\_LOAN}$\rightarrow$\textsc{DE\_LOAN}: 18{,}142 cases across all model and prompt combinations), which is 4.6$\times$ more frequent than the reverse direction
(\textsc{DE\_LOAN}$\rightarrow$\textsc{FR\_LOAN}: 3{,}946). Native Luxembourgish items are also misattributed to German borrowings (20{,}662) substantially more often than to French borrowings (5{,}944), indicating an overall tendency to overpredict \textsc{DE\_LOAN}. This pattern is consistent with potential lexical/orthographic overlap between French- and German-origin forms in Luxembourgish, although other factors (e.g., class priors or KG coverage) may also contribute. Overall, these results support RQ2: while KG-graph improves borrowing recognition, donor identification remains skewed toward German across model and prompt settings.

\paragraph{EN\_LOAN as a distractor label.}
Although EN\_LOAN is absent from the evaluation set (only 24 source instances, below the 50-instance threshold), we retain it as a valid output label to probe whether models project English-origin hypotheses onto tokens that are in fact native or borrowed from French or German. Table~\ref{tab:en_distractor} reports how often each model predicts EN\_LOAN and which true class absorbs those false positives.

\begin{table}[hpt!]
\centering
\scriptsize
\setlength{\tabcolsep}{4pt}
\renewcommand{\arraystretch}{0.45}
\caption{EN\_LOAN false-positive analysis.\ \textbf{EN\textsubscript{pred}} is the
total number of EN\_LOAN predictions; columns show the true-class
breakdown of those predictions.\ \textbf{Rate} is the proportion of all
valid predictions assigned to EN\_LOAN.}
\label{tab:en_distractor}
\begin{tabular}{l l r r r r r}
\toprule
\textbf{Model} & \textbf{Prompt} & \textbf{EN\textsubscript{pred}}
  & \textbf{$\rightarrow$NAT} & \textbf{$\rightarrow$FR} & \textbf{$\rightarrow$DE}
  & \textbf{Rate} \\
\midrule
Gemma~12B  & Zero-shot & 1\,001 & 392 & 196 & 387 & 32.8\% \\
           & Few-shot  &   238  & 103 &  29 &  97 &  7.8\% \\
           & KG-flat   &   627  & 254 &  80 & 275 & 20.6\% \\
\rowcolor{kgrow}
           & KG-graph  &   128  & 107 &   5 &  12 &  4.2\% \\
\midrule
Gemma~27B  & Zero-shot &   312  & 130 &  47 & 129 & 10.2\% \\
           & Few-shot  &   212  &  96 &  23 &  89 &  7.0\% \\
           & KG-flat   &   299  & 125 &  31 & 136 &  9.8\% \\
\rowcolor{kgrow}
           & KG-graph  &   167  & 128 &   6 &  28 &  5.5\% \\
\midrule
Llama~70B  & Zero-shot &   363  & 196 &  37 & 125 & 12.0\% \\
           & Few-shot  &    91  &  38 &   6 &  43 &  3.0\% \\
           & KG-flat   &    87  &  32 &   7 &  45 &  2.9\% \\
\rowcolor{kgrow}
           & KG-graph  &   264  & 222 &   7 &  32 &  8.7\% \\
\bottomrule
\end{tabular}
\end{table}

Two patterns stand out.
First, without structured context, models frequently over-predict EN\_LOAN: Gemma~12B assigns it to nearly a third of all tokens under zero-shot prompting.
KG-graph prompting reduces the EN\_LOAN rate by 78--87\% for the Gemma models (from 32.8\% to 4.2\% for Gemma~12B, and from 10.2\% to 5.5\% for Gemma~27B), confirming that structured linguistic context suppresses spurious English-origin hypotheses.
Second, across all models and prompt conditions, the majority of false EN\_LOAN predictions fall on genuinely \textsc{NATIVE} tokens rather than on French or German borrowings.
Under KG-graph, this concentration intensifies: 84\% of Gemma~12B's and 77\% of Gemma~27B's residual EN\_LOAN predictions fall on \textsc{NATIVE} tokens.
This suggests that when models lack donor-specific evidence, they default to an English-origin hypothesis for unfamiliar Luxembourgish words, a bias consistent with English's dominance in multilingual pre-training corpora.

Retaining EN\_LOAN as a distractor label therefore serves a diagnostic purpose: it exposes this bias and provides a measurable signal of how effectively structured context can counteract it.

\subsection{RQ3. Ablating KG components}
\label{subsec:rq3}
\begin{figure}[hpt!]
  \centering
  \includegraphics[width=\linewidth]{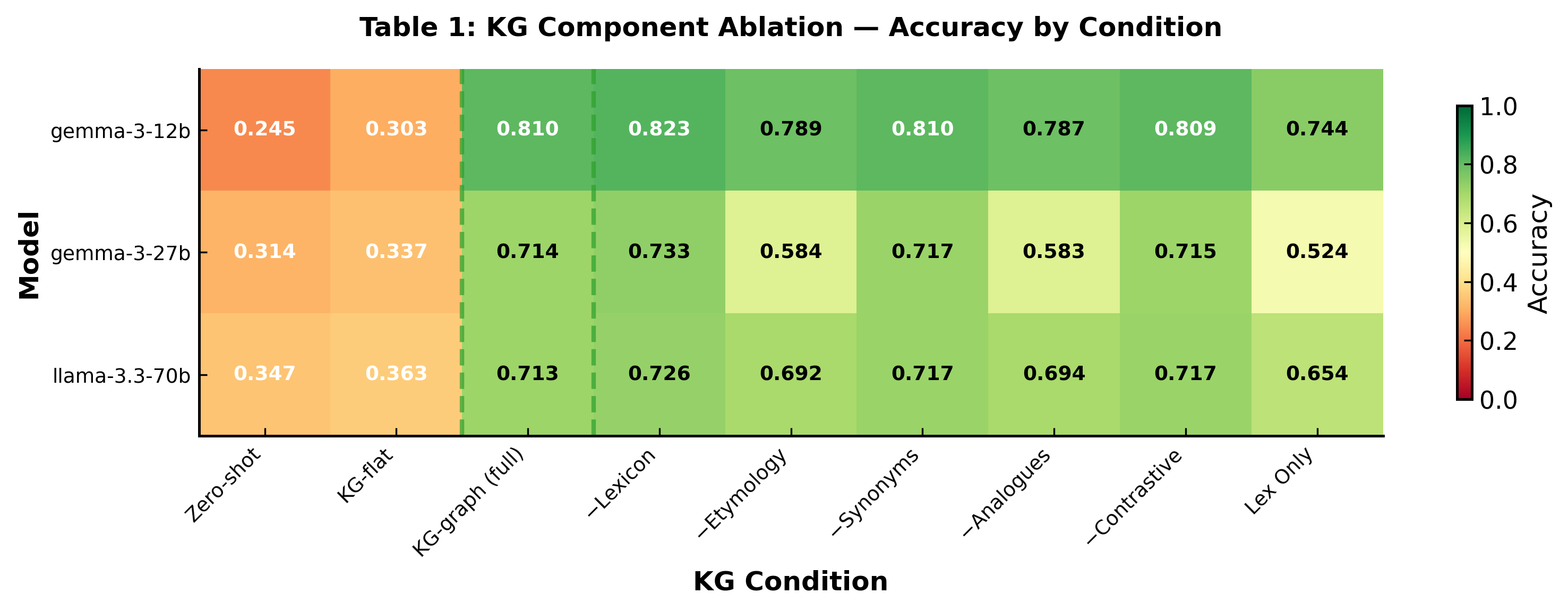}
  \caption{Impact of KG components on borrowing accuracy by model and KG-graph ablation condition.}
  \label{fig:ablation_kg_component}
\end{figure}

Figure~\ref{fig:ablation_kg_component} shows the effect of removing individual components from the KG-graph prompt. 
The full KG-graph condition reaches 81.0\%, 71.4\%, and 71.3\%
accuracy for the three models (Gemma~3~12B, Gemma~3~27B, and
Llama~3.3~70B).
Removing etymological information (No Etymology) reduces accuracy to
78.9\% for Gemma~3~12B, 58.4\% for Gemma~3~27B, and 69.2\% for
Llama~3.3~70B.
Dropping analogical examples (No Analogues) has a similarly strong
impact, especially on the 27B model, where accuracy decreases by
roughly 13 percentage points.

By contrast, removing synonym links or contrastive patterns changes
performance only marginally, within $\pm$0.3 points of the full
KG-graph condition.
A Lexicon-only variant that keeps dictionary entries but discards graph structure clearly outperforms non-KG baselines, yet remains 6–19 points behind the full graph, which suggests that donor chains and pattern-sharing analogues carry most of the useful signal, while long definitions may introduce noise.
In some settings, accuracy even improves slightly when the lexicon text is
removed, but the graph structure is kept, reinforcing that relational structure is more valuable than raw definitional prose.


\subsection{Model scale and benefit from KG}
\label{subsec:scale}



Table~\ref{tab:overall_performance} compares zero-shot and KG-graph accuracy by model size. Zero-shot accuracy grows modestly with scale, from 24.5\% for Gemma~3~12B to 34.7\% for Llama~3.3~70B, but under KG-graph the ranking inverts: Gemma~3~12B reaches 81.0\%, while Gemma~3~27B and Llama~3.3~70B plateau at 71.4\% and 71.3\%, respectively. The KG gain $\Delta_{\mathrm{KG}}$, defined as the accuracy difference between KG-graph and zero-shot, decreases monotonically with model size: +56.5, +40.1, and +36.6 percentage points. A supplementary log-scale visualization of this trend is provided in Appendix~\ref{fig:scale_kg_gain}.

\paragraph{The reversal is KG-specific.}
Gemma~12B does \emph{not} generally outperform Gemma~27B. Under zero-shot (31.4\% vs.\ 24.5\%), few-shot (38.0\% vs.\ 38.3\%, essentially tied), and KG-flat (33.7\% vs.\ 30.3\%), Gemma~27B matches or exceeds Gemma~12B. The reversal occurs exclusively under KG-graph (+9.5~pp in favor of 12B), ruling out a general quality advantage of the smaller model and localizing the effect to how each model utilizes instance-specific structured context.

\paragraph{Mechanism: NATIVE over-prediction by larger models.}
Under KG-graph, all models achieve high recall on \textsc{FR\_LOAN} ($\geq$0.860) and \textsc{DE\_LOAN} ($\geq$0.961), but \textsc{NATIVE} recall drops sharply with scale: 0.639 (12B), 0.447 (27B), 0.349 (70B). Precision on \textsc{NATIVE} remains above 0.94 for all models, so larger models predict \textsc{NATIVE} correctly when they do---but they predict it far too rarely, over-attributing borrowing status to native words. KG ablations confirm this asymmetry: removing etymology or analogues costs Gemma~27B ${\sim}$13~pp but Gemma~12B only ${\sim}$2~pp, showing that the larger model falls back on parametric borrowing priors when graph evidence is incomplete.

This pattern is consistent with findings on parametric--contextual knowledge conflicts~\cite{longpre2021entity,xie2023adaptive}: larger models develop stronger internal representations of French and German items during pre-training, which compete with KG-supplied evidence and lead to over-attribution of donor origins. The smaller model, lacking such entrenched priors, defers more faithfully to the structured context.

In this analysis, we use publication dates as a diachronic proxy to contrast more established items with more recent adaptations (pre-2015 vs.\ post-2015).
The graph encodes structural origin information (donor language, morphological patterns, analogues) but not explicit recency cues such as frequency trajectories or first-attestation dates. Under this temporal split (see Supplementary), KG-graph is the most temporally robust condition, with recent vs.\ established gaps of only 0.7--2.8~pp.

Under the KG-graph condition, Gemma~3~12B achieves 81.4\% accuracy
for established items and 80.7\% for recent ones; Gemma~3~27B
achieves 73.1\% and 70.3\%; Llama~3.3~70B reaches 72.4\% and 70.6\%.
These results indicate that the graph captures structural regularities that transfer to more recent lexical items, even if such items are under-represented or missing in the models’ pre-training data.
Among all prompt strategies, KG-graph is the least affected by recency, suggesting that structured linguistic context can partially compensate for gaps in parametric training data; a full breakdown by model and prompt is provided in the supplementary material.

\subsection{Neology detection}
\label{subsec:neology}

The neology decision task, which collapses all borrowings into a single lexical-innovation class versus native items, behaves very differently from borrowing classification.
Table~\ref{tab:neology} reports accuracy and $F1_{\text{neo}}$ for
the ``neologism'' class by model and prompt strategy. Here, few-shot prompting is consistently the most effective strategy.
Gemma~3~12B reaches 48.5\% accuracy and
$F1_{\text{neo}} = 0.509$, Gemma~3~27B reaches 49.2\% and 0.524,
and Llama~3.3~70B achieves 40.8\% and 0.254.
In contrast, the KG-graph condition substantially degrades performance.
Accuracy falls to 34.2\%, 30.3\%, and 30.5\% for the three models,
and $F1_{\text{neo}}$ for Llama~3.3~70B drops close to zero.

This divergence is in line with how the linguistic knowledge graph is
constructed.
The graph encodes origin and structural information (donor language,
morphological pattern, analogues, native synonyms), which are exactly
the cues needed for borrowing classification, but largely orthogonal to
\emph{recency}.
Deciding whether a word counts as a lexical innovation requires diachronic evidence, such as frequency trajectories, first attestation dates, or domain-specific usage shifts, none of which are currently exposed in the graph.
As a result, the additional context encourages models to reason about
\emph{where} a word comes from rather than \emph{when} it entered the
language, which can mislead them in borderline cases.

\begin{table}[hpt!]
  \centering
  \scriptsize
  \setlength{\tabcolsep}{12pt}
  \renewcommand{\arraystretch}{0.45}
  \caption{Neology decision performance by model and prompt.
  Accuracy and $F1_{\text{neo}}$ for the ``neologism'' class.}
  \label{tab:neology}
  \begin{tabular}{l l r r}
    \toprule
    \textbf{Model}         & \textbf{Prompt}    & \textbf{Acc. (\%)} & $F1_{\text{neo}}$ \\
    \midrule
    Gemma~3~12B   & Zero-shot & 41.2      & 0.308             \\
    \rowcolor{kgrow}
                  & Few-shot  & 48.5      & 0.509             \\
                  & KG-graph  & 34.2      & 0.042             \\
    \midrule
    Gemma~3~27B   & Zero-shot & 45.6      & 0.429             \\
    \rowcolor{kgrow}
                  & Few-shot  & 49.2      & 0.524             \\
                  & KG-graph  & 30.3      & 0.064             \\
    \midrule
    Llama~3.3~70B & Zero-shot & 36.7      & 0.127             \\
    \rowcolor{kgrow}
                  & Few-shot  & 40.8      & 0.254             \\
                  & KG-graph  & 30.5      & 0.012             \\
    \bottomrule
  \end{tabular}
\end{table}

\subsection{Binary native versus borrowed}
\label{subsec:binary}

Finally, we collapse the four-class label space into a binary decision
and ask models to distinguish native Luxembourgish words from any type
of borrowing.
Under the KG-graph condition, all models reach high performance.
Gemma~3~12B attains 85.5\% accuracy and $F1 = 0.902$,
Gemma~3~27B reaches 78.9\% and 0.862, and Llama~3.3~70B reaches
75.5\% and 0.845.

The contrast between the binary decision and the donor-specific four-way task suggests that the main residual difficulty lies in
separating French from German borrowings, rather than in detecting
whether a token is lexically integrated at all.
In other words, once the knowledge graph is available, knowing that a word is a borrowing is comparatively easy, while pinpointing the
correct donor in a dense Luxembourgish, French, and German contact zone
remains challenging.
Detailed binary results for all prompt strategies are reported in
the supplementary material.

\section{Discussion}
\label{sec:discussion}

Our results show that off-the-shelf multilingual LLMs have limited awareness of how a small contact language integrates lexical borrowings, even when trained on large multilingual corpora.
With four possible output labels, a random baseline yields 25\% accuracy;
zero-shot performance ranges from 24.5\% to 34.7\%, indicating that parametric knowledge alone barely exceeds chance. This observation aligns with work in contact linguistics and neology that emphasizes community entrenchment, dictionary listedness, and usage patterns over purely formal cues \citep{treffersdaller2025simple,chesley2010predicting,wolfer2023tracking}. The models do not spontaneously replicate the ``Simple View'' of borrowing as operationalized in lexicographic and corpus studies.


LexNeo-Bench complements earlier borrowing and anglicism corpora in Spanish and other languages \citep{alvarezmellado2020annotated,alvarez2021extracting,adobo2021,adobo2025,alvarezmellado2020lazaro,kevers2022coswid} by exposing LLMs to a dense Luxembourgish, French, and German contact zone where orthographic and morphological integration is pervasive \citep{adda2008developments,lavergne2014automatic,anastasiou2022deliverable}. The strong gains from structured knowledge-graph prompting suggest that models can make fine-grained borrowing decisions once they are supplied with token-specific morphological patterns, donor labels, and analogical examples. This mirrors gains observed when injecting gazetteers and knowledge bases into NER and entity-centric tasks \citep{tan2023damo,chen2022ustc} and supports the view that community lexical resources remain crucial even in the LLM era \citep{tomaszewska2025neon,hosseini2025hybrid}.

At the same time, our neology decision results highlight that structural donor information alone does not solve diachronic questions. LLMs perform best with few-shot prompting that clarifies the task
mapping (borrowings count as lexical innovations), while knowledge-graph prompts, which were designed for borrowing classification, can even harm performance. This gap reflects broader findings on LLM-based neology detection and ``LLM neologisms'' that arise from tokenization and encoding artifacts rather than organic community usage \citep{iwamoto2024llm,zheng2024neo}. For small languages with sparse written production and heavy code mixing \citep{plum2024luxbank,adda2008developments}, separating genuine innovations from long-standing borrowings remains challenging without explicit temporal signals or external diachronic corpora.

Our study has several limitations: First, LexNeo-Bench is derived from a single edited news source, so it under-represents informal registers and spoken discourse. Second, borrowing labels rely on an automatic pattern pipeline and dictionary signals, which may misclassify borderline items or miss emerging forms in under-documented domains. Third, we evaluate only three instruction-tuned models with frozen prompts, so conclusions about model scale and architecture should be treated as tentative. Finally, the benchmark focuses on token-level decisions and does not directly measure how LLMs handle borrowing in generation, for example, in spelling correction or style transfer. Addressing these limitations will require extending the benchmark to other genres, adding human validation for difficult cases, and coupling classification with controlled generation tasks.

\section{Conclusion and Future Work}
\label{sec:conclusion}
We introduced LexNeo-Bench, a token-level benchmark derived from a borrowing-annotated Luxembourgish news corpus to probe how multilingual LLMs treat morphologically adapted borrowings. Across three models and 34 prompt configurations, zero-shot parametric knowledge stays near chance, whereas instance-specific linguistic knowledge graphs raise borrowing
classification accuracy to about 71--81\% and substantially improve binary native versus borrowed decisions, with the largest gains for the smallest model. This shows that structured lexical context can partly compensate for sparse pretraining in low-resource contact languages, while neology decisions remain difficult and are best supported by few-shot prompting in our experiments because recency is not encoded in the current graph. Future work will add explicit diachronic signals, extend LexNeo-Bench beyond edited news and Luxembourgish, and link token-level evaluation to downstream writing assistance to quantify the user-facing impact of borrowing misclassifications.

\section{Acknowledgements}
We thank RTL Luxembourg and Tom Weber for providing access to the news archive and for supporting its use for research purposes. This work highly benefited from the collaborative network fostered by the \textbf{ENEOLI COST Action (CA22126)}, supported by COST (European Cooperation in Science and Technology), and also within the project LuxVoice (project reference 19205922) from the FNR. 

\section{Ethical and legal aspects}
\label{sec:ethics}

\paragraph{Data provenance and legal basis.}
The underlying corpus consists of online news articles published between 1999 and 2025 by a major Luxembourgish media outlet (RTL). The data were obtained under a formal research collaboration and processed under the outlet’s terms of use and the applicable EU text and data mining provisions for non-commercial scientific research. No user accounts were accessed, no technical protection measures were circumvented, and we did not perform large-scale scraping of the public-facing website.

\paragraph{Data and code availability.}
Full prompt templates for all strategies and tasks, including the complete five-example few-shot prompt and the neology template, will be provided in the public GitHub: ~\texttt{github.com/NinaKivanani/LexNeo-Bench}.

\paragraph{Intellectual property and data release.}
All source articles remain under the copyright and database rights of RTL. Our preprocessing, annotation, and analysis operate on copies stored on secure institutional infrastructure; we do not redistribute the full text of the corpus. Instead, we release only derived artifacts that are not substitutable for the original content, including the annotation schema and pattern inventory, scripts to reproduce the pipeline on any legally obtained Luxembourgish news corpus, aggregate statistics and plots, and small illustrative excerpts.

\paragraph{Privacy and data protection.}
News articles naturally contain references to identifiable individuals. These mentions appear in material already made lawfully available online in the exercise of journalistic freedom, yet they still qualify as personal data. We do not link the corpus to external records, attempt to profile individuals, or infer sensitive attributes. All analyses are conducted at the token, sentence, or aggregate document level rather than at the level of specific persons. Data are stored and processed on secure servers, including national high-performance computing resources, in compliance with the GDPR and relevant national data protection requirements.

\paragraph{Intended use and potential impact.}
LexNeo-Bench reflects the editorial practices and topic mix of a single news provider and should not be treated as representative of all Luxembourgish language use. The benchmark and LKG are intended as descriptive tools for studying contact phenomena and for stress-testing multilingual LLMs on borrowing and neology, not as prescriptive standards for ``correct'' Luxembourgish. We explicitly discourage using these resources to police lexical borrowing, to stigmatize code-switching in everyday communication, or to draw strong sociolinguistic conclusions about specific groups. Any correlations between borrowing patterns and social or regional factors must be interpreted with caution to avoid reinforcing stereotypes or over-generalising from a single, institutionally edited source.


\section{Bibliographical References}\label{sec:reference}

\bibliographystyle{lrec2026-natbib}
\bibliography{lrec2026-example}

\section{Language Resource References}
\label{lr:ref}
\bibliographystylelanguageresource{lrec2026-natbib}
\bibliographylanguageresource{languageresource}

\section{Appendices}
\subsection{Supplementary visualization of KG gain}
KG gain is defined as the accuracy difference between KG-graph and zero-shot prompting. The gain decreases monotonically with scale, from +56.5 percentage points for Gemma~3~12B to +40.1 for Gemma~3~27B and +36.6 for Llama~3.3~70B.
\begin{figure*}
    \centering
    \includegraphics[width=0.8\textwidth]{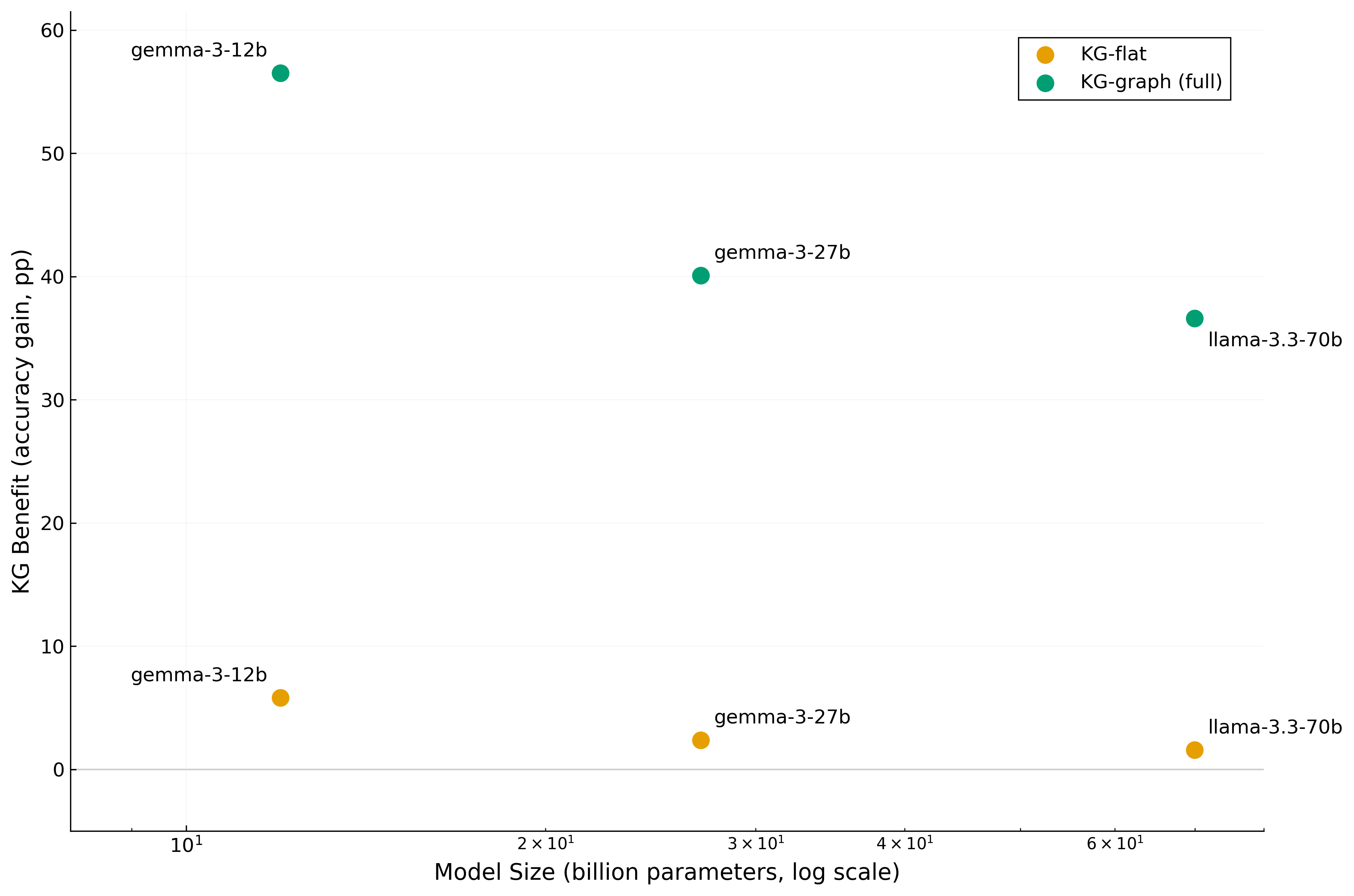}
    \caption{Supplementary visualization of KG gain $\Delta_{\mathrm{KG}}$ by model size on a log-scaled x-axis.}
    \label{fig:scale_kg_gain}
\end{figure*}

\end{document}